\documentclass[pdflatex,sn-mathphys-num,referee]{sn-jnl}

\usepackage{bm}
\usepackage{pifont}
\usepackage{adjustbox}
\usepackage[dvipsnames]{xcolor}
\usepackage{graphicx}%
\usepackage{multirow}%
\usepackage{amsmath,amssymb,amsfonts}%
\usepackage{amsthm}%
\usepackage{mathrsfs}%
\usepackage[title]{appendix}%
\usepackage{xcolor}%
\usepackage{textcomp}%
\usepackage{manyfoot}%
\usepackage{booktabs}%
\usepackage{algorithm}%
\usepackage{algorithmicx}%
\usepackage{algpseudocode}%
\usepackage{listings}%
\usepackage{xcolor}
\usepackage{soul}
\sethlcolor{yellow}

\theoremstyle{thmstyleone}%
\theoremstyle{thmstyletwo}%

\theoremstyle{thmstylethree}%

\newcommand*\colourcheck[1]{%
  \expandafter\newcommand\csname #1check\endcsname{\textcolor{#1}{\ding{52}}}%
}
\newcommand*\colourcheckodd[1]{%
  \expandafter\newcommand\csname #1checkodd\endcsname{\textcolor{#1}{\ding{51}}}%
}
\newcommand*\colourxmark[1]{%
  \expandafter\newcommand\csname #1xmark\endcsname{\textcolor{#1}{\ding{54}}}%
}
\definecolor{bloodred}{HTML}{B00000}
\definecolor{cautionyellow}{HTML}{EED202}

\colourcheckodd{cautionyellow}
\colourcheck{cautionyellow}
\colourcheck{OliveGreen}
\colourxmark{bloodred}

\begin{document}

\title[Article Title]{A Scalable Embodied Intelligence Platform for Seamless Real-to-Sim-to-Real Transfer of Household Mobile Manipulation Tasks}


\author[1]{\fnm{Kui} \sur{Yang}}\email{yangkui1127@gmail.com}
\equalcont{These authors contributed equally to this work.}

\author[1]{\fnm{Xianlei} \sur{Long}}\email{xianlei.long@cqu.edu.cn}
\equalcont{These authors contributed equally to this work.}

\author[1]{\fnm{Haoxuan} \sur{Li}}\email{20214707@cqu.edu.cn}



\author*[3]{\fnm{Yan} \sur{Ding}}\email{yding25@binghamton.edu}

\author*[1]{\fnm{Chao} \sur{Chen}}\email{cschaochen@cqu.edu.cn}

\affil[1]{\orgdiv{School of Computer Science}, \orgname{Chongqing University}, \orgaddress{\street{No. 55 Daxuecheng South Road, Shapingba District}, \city{Chongqing}, \postcode{401331}, \state{Chongqing}, \country{China}}}


\affil[3]{\orgdiv{R\&D Department}, \orgname{Lumos Robotics Technology (Suzhou) Co., Ltd}, \orgaddress{\street{No. 345 Baodai East Road, Wuzhong District}, \city{Suzhou}, \postcode{215128}, \state{Jiangsu}, \country{China}}}


\abstract{Mobile manipulation is a fundamental capability in embodied intelligence robotics. The growing demand for robust and generalizable manipulation in unstructured household environments has driven rapid progress in embodied intelligence platforms. However, achieving a seamless transfer across the real-to-sim-to-real cycle faces three key challenges, including costly high-fidelity simulation scenes reconstruction, the complexity of systematic strategy evaluation in simulation, and incompatible real-world deployments. To address these challenges, we develop \textbf{BestMan}, a scalable and seamless real-to-sim-to-real platform that bridges the gap between the simulation and the real world, enabling effective strategy development, integration, and deployment for household mobile manipulation. Specifically, we design a novel Automated Scene Generation (ASG) module to reconstruct realistic \textit{simulations} from \textit{real} observations. Then, we propose a simulation-guided task formalization and skill learning architecture that supports the flexible integration and large-scale evaluations of hybrid skill strategies in \textit{simulation}. Finally, to enhance the real-world scalability, we develop a Hardware-agnostic and Unified Middleware (HUM) to ensure seamless and compatible sim-to-real transfer across heterogeneous mobile manipulators for \textit{real} deployments. Experimental results demonstrate the superior performance of our proposed platform in establishing standardized benchmarks and facilitating promising research in the field of mobile manipulation.}


\maketitle

\section{Introduction}\label{sec1}

Embodied Intelligence (EI) has emerged as a promising paradigm for developing autonomous systems that perceive, reason, and act within the real world~\cite{duan2022survey, liu2024survey, tian2025assisting, Acbot}. Among various embodied agents, mobile manipulators are representatives, combining mobility and dexterous manipulation to perform complex interactions in diverse settings~\cite{thakar2023survey, LG-DSG}. In everyday household environments such as kitchens, mobile manipulation enables agents to follow natural language instructions, e.g., \textit{``Pick up the Cola from the refrigerator and place it inside the cabinet''}.
 Robustly executing such tasks under diverse real-world scenarios requires the integration of different functional systems~\cite{liu2025re}, including coordinate perception~\cite{OVA-Fields}, task planning~\cite{SKE-Layout}, motion control~\cite{moma_pos}, and object manipulation across multiple stages.
However, deploying and testing these systems directly on physical robots remains costly and inefficient, due to hardware costs, lengthy experimentation cycles, safety concerns, and poor generalization across scenes~\cite{liu2024survey, zhang2025global}. These limitations highlight the critical role of simulation in developing and scaling mobile manipulation capabilities.

Recent research on EI platforms for mobile manipulation is in a stage of rapid growth. Several general-purpose simulators have been developed~\cite{gazebo, mujoco, sapien}, serving as the foundation for the task-oriented simulation framework building~\cite{virtualhome, habitat, robocasa, igibson}. While these platforms have accelerated algorithmic development in simulation, they often rely on manual or procedural scene generation and lack an automated reconstruction pipeline for accurately reflecting real-world disturbances to improve robustness, which underscores the importance of real-to-sim transfer. In contrast, existing real-world platforms~\cite{homerobot, ok-robot} are typically built on the Robot Operating System (ROS)~\cite{ros} to explore the deployment performance. However, they are often tightly coupled with specific hardware, making it difficult to connect simulations for rapid, large-scale testing and requiring huge reconfigurations and transfer to heterogeneous robotic hardware~\cite{CL-OVMM-GPT4V}. Although some platforms attempt to integrate both simulation and real-world deployments~\cite{orbit, robosuite}, they are often tailored to specific tasks, lacking the generality and scalability required for the complex mobile manipulation. In summary, these limitations reveal a critical need for a scalable real-to-sim-to-real platform that seamlessly bridges scene reconstruction, strategy learning, and real-world deployments.

Despite its promise, building a  {scalable and seamless} platform for household mobile manipulation in unstructured environments faces several key challenges. 
First, accurate and scalable real-to-sim scene reconstruction remains a bottleneck. Manually constructing simulation environments using scanned 3D assets~\cite{google_scanned_dataset, kitchen_worlds} is labor-intensive and not scalable, limiting the quality and quantity of data usable for training. Second, the long-horizon nature of household tasks introduces significant difficulties in perception, planning, and control~\cite{Robovqa, E2VPT, AMD}. These tasks involve multi-stage objectives, cluttered environments, and fine-grained object interactions, but lack standardized decomposition and evaluation protocols. 
Third, achieving reliable and seamless sim-to-real deployments is hindered by mismatches between simulation frameworks and real-world hardware interfaces, as well as by the diversity of robot platforms and control APIs~\cite{MF-S2R}. These factors make it difficult to deploy learned policies across diverse real-world platforms, limiting the reusability and scalability of mobile manipulation systems.


\begin{figure}[t]
    \centering
    \includegraphics[width=\textwidth]{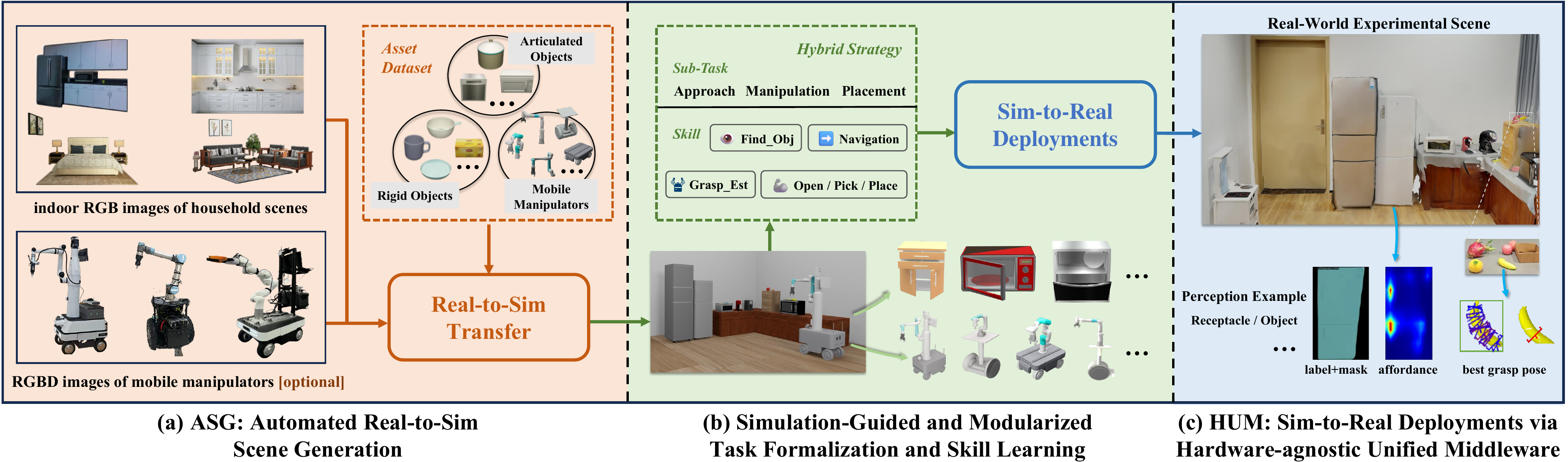}
    \caption{Overview of the developed \textbf{BestMan} (Real-to-Sim-to-Real) platform, consisting of three primary modules: (a) ASG for real-to-sim transfer; (b) simulation-guided and modularized task formalization and skill learning; (c) HUM for sim-to-real deployments. Together, these modules facilitate \textbf{BestMan}'s scalable algorithmic developments and seamless real-world deployments for complex household mobile manipulation tasks and environments.}
    \label{fig:overview}
\end{figure}

To address above-mentioned challenges, in this study, we propose and develop \textbf{BestMan}, a {\em scalable and seamless} mobile manipulation platform that integrates real-to-sim scene reconstructions,  algorithm developments, and deployments across heterogeneous platforms. Specifically, to overcome limitations of the manual simulation setup, we propose an Automated real-to-sim Scene Generation (ASG) module that reconstructs articulated and semantically aligned simulation scenes directly from real-world images. Second, to support systematic algorithm development, we establish the Transferable Household Mobile Manipulation (THMM) benchmark, which leverages a modular Sense-Plan-Act (SPA) architecture to decompose long-horizon tasks into reusable perception and planning skills, enabling standardized task formalization and evaluation. Then, to ensure wide hardware compatibility, a Hardware-agnostic Unified Middleware (HUM) that bridges simulations and real-world systems is proposed, supporting scalable sim-to-real deployments across heterogeneous robots. Finally, extensive experimental results verify the scalability and seamlessness of the developed platform in several typical household mobile manipulation tasks.

We substantially extends our preliminary short-paper work\mbox{~\cite{short_version}}, which introduced a foundational, simulation-centric modular architecture. To address its limitations—specifically the absence of formalized evaluation benchmark and real-world deployment capabilities—the present study establishes a comprehensive real-to-sim-to-real pipeline. Our novel contributions are explicitly summarized as follows:

\begin{itemize}
\item We develop a scalable and seamless real-to-sim-to-real platform, \textbf{BestMan}, which is a unified and extensible platform for the  mobile manipulation that integrates real-world scene understanding, skill-based task formalization, and sim-to-real transfer capabilities, supporting the end-to-end deployments of task-specific algorithms. 
\item  We develop the ASG for real-to-sim transfer, which reconstructs articulated and semantically grounded simulation scenes directly from real-world images, enabling data-efficient training and evaluation.
\item We build the THMM benchmark in a modular integration manner, which formalizes long-horizon mobile manipulation tasks using a Sense-Plan-Act structure and supports reusable skill modules for perception and planning.
\item We design the HUM for sim-to-real deployments that bridge simulation and real systems, enabling scalable and compatible operation on heterogeneous robots.
\end{itemize}

\section{The BestMan platform}\label{sec2}

\subsection{Overview}
An overview of the developed \textbf{BestMan} platform is illustrated in Figure~\ref{fig:overview}. Our goal is to enable seamless real-to-sim-to-real transfer for mobile manipulation tasks.
It consists of three primary modules:
(1) ASG for real-to-sim transfer, as shown in Figure~\ref{fig:overview}-(a), where geometrically and semantically realistic simulation environments are automatically reconstructed from real-world images;
(2) Simulation-guided and modularized task formalization and skill learning is shown in Figure~\ref{fig:overview}-(b), where household mobile manipulation tasks are formalized as language-guided tasks, decomposed into sequential sub-tasks, and solved using a modularized skill architecture;
(3) HUM for sim-to-real deployments, as illustrated in Figure~\ref{fig:overview}-(c), where optimized hybrid strategies are transferred to the real world, enabling robust execution across heterogeneous  mobile manipulators.

To achieve this, we integrate a high-fidelity simulation framework using PyBullet for physics~\cite{pybullet} and Blender for rendering~\cite{blender}, together with a unified real-world robotic system that supports deployments across heterogeneous hardware. 

\begin{figure}[t]
    \centering
    \includegraphics[width=\textwidth]{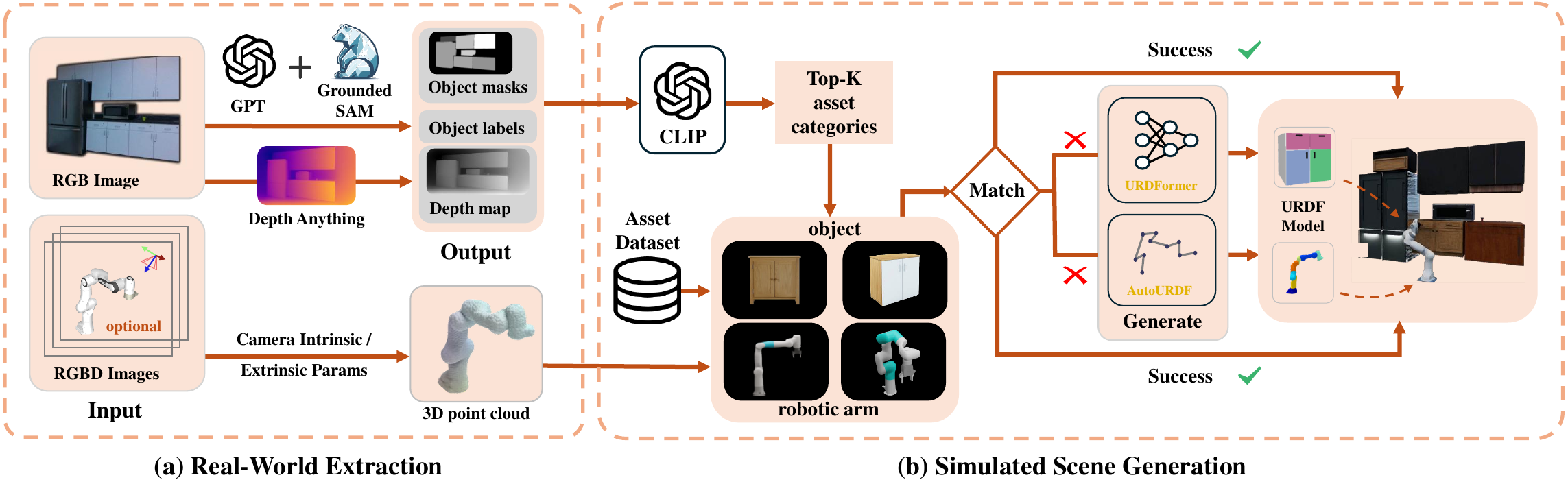}
    \caption{Overview of the ASG, which comprises  two steps: (a) Real-World Perception and Extraction, which extracts relevant object-level information using off-the-shelf visual models, along with constructing robot point clouds based on camera parameters; (b) Simulated Scene and Asset Generation, in which each detected object is matched to a digital asset from the asset dataset based on the semantic and geometric similarity. If no suitable match is claimed, it will be generated by the URDFormer model. Simultaneously, the robot's URDF model is constructed from point clouds using AutoURDF. All assets are then integrated into a physically consistent simulated scene.}
    \label{fig:real-to-sim}
\end{figure}

\subsection{ASG: Automated real-to-sim scene generation}
To address the challenge of accurately extracting and recovering semantic information from raw real-world observations, we propose ASG to construct geometrically and semantically realistic simulated scenes. Our real-to-sim pipeline is designed at the part-level granularity, enabling fully-interactive and physically-plausible simulations, rather than generating a single globally-unified geometry or a locally manipulable subset. To achieve this, it requires (1) the extraction of accurate semantic information from real-world images of indoor environments, and (2) the generation of interactive joint structures and physical parameters for the objects. To this end, we break down the real-to-sim process into two steps: the first  is a real-world \textit{extraction} step, in which  relevant objects and the robot information are extracted from the raw camera image; and the second  is a digital asset \textit{generation} step, in which URDF models are generated to form a complete digital scene. An overview of the ASG is provided in Figure~\ref{fig:real-to-sim}.

\subsubsection{Real-World perception and extraction} 
The first step requires a single-view RGB image $X_1$ of the environment and multi-view whole-body RGBD images $X_2$ of the robot, captured by a calibrated camera with intrinsic matrix $K$. To extract the object masks for $X_1$, we first prompt GPT-4 to retrieve all object labels $L_i = \{l_1, l_2, ..., l_i\}, i \in \{1,2,...,n\}$. Then labels are passed to Grounded-SAM-v2~\cite{grounded_sam} as text prompts to generate object masks $M = \{m_1, m_2, ..., m_i\}, i \in \{1,2,...,n\}$. Due to the potential degradation of the depth camera performance caused by reflections in the scene, we use the Depth-Anything-v2~\cite{depth_anything_v2} model to extract the depth map $D$ from $X_1$. We then obtain the scene point cloud $P$ and leverage the individual object mask $m_i$ to generate a point subset $P_i$. Simultaneously, we perform point cloud fusion to extract the complete robot point cloud from multi-view RGBD images, which is optional since more accurate models can be obtained from the manufacturer and is only used when the full configuration is unavailable.

\subsubsection{Simulated scene generation} 
Based on the previously extracted object point cloud $P_i$, we perform hierarchical retrieval from the 3D simulation asset dataset by selecting the top $k$ most similar categories, determined by the CLIP score~\cite{clip} between the object label $l_i$ and all asset categories. Given the selected categories, the closest digital asset is identified by computing the DINOv2 feature embedding distance~\cite{dino_v2} between the object $x_i$ and each candidate asset. If the computed distance exceeds a predefined threshold, indicating the absence of a functionally similar object, we employ URDFormer~\cite{urdformer} to model the object's accurate articulated structure. Similarly, if no suitable robot model is available, AutoURDF~\cite{autourdf} is employed to construct its articulated structure. Finally, the generated assets are integrated into a simulation scene, aligned with the extracted spatial information to ensure physical consistency. The final simulation scene is shown in the right part of Figure \ref{fig:real-to-sim}.

\begin{figure}[t]
\centering
\includegraphics[width=\columnwidth]{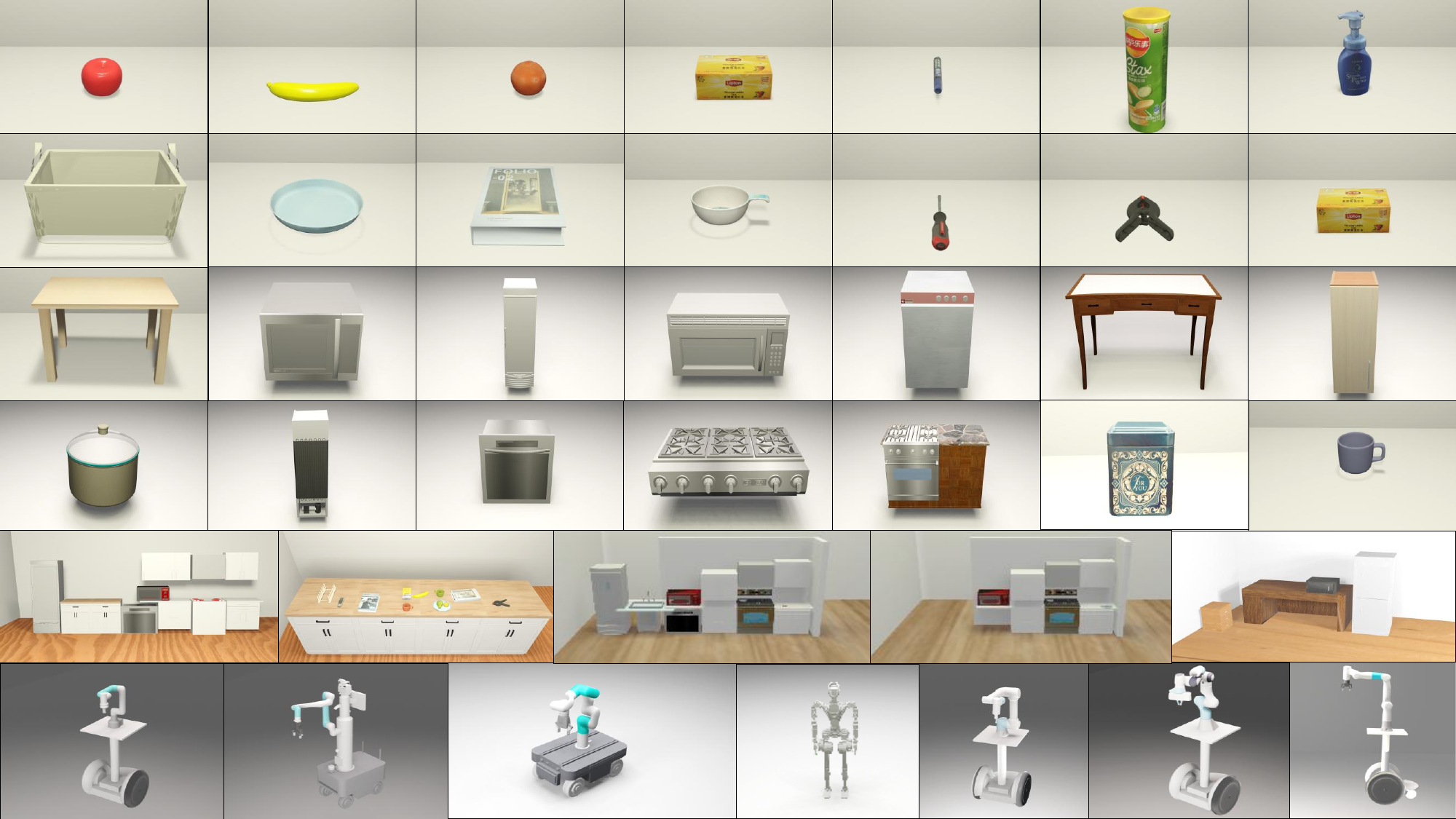}
\caption{Overview of the asset dataset. From top to bottom, encompasses rigid objects, articulated objects, part-level manipulable environments, and various modular mobile manipulators.}\label{fig:asset}
\end{figure}

\subsubsection{Asset dataset}
As illustrated in Figure~\ref{fig:asset}, our asset library comprises high-fidelity object models and modular robotic components designed for seamless real-to-sim integration. To ensure diverse and realistic interactions, we incorporate several large-scale scanned datasets, including rigid objects from YCB~\cite{ycb_dataset}, IPA-3D1K~\cite{ipa_3d1k}, and Google Scanned Objects~\cite{google_scanned_dataset}, alongside articulated models from PartNet-Mobility~\cite{partnet}. These datasets encompass a wide array of categories—ranging from small tools to household appliances—providing high-quality, multi-view 3D geometries and part-level kinematic information. This extensive repository enables our pipeline to generate simulated scenes that accurately reflect real-world geometry and articulation, thereby facilitating effective manipulation training and validation.

Beyond static objects, we have pre-integrated a comprehensive suite of Unified Robot Description Format (URDF) models to support the synthesis of mobile manipulators. These models are designed with a modular architecture, allowing for rapid hardware assembly and configuration without requiring structural modifications. The robot assets are categorized as follows:
\begin{itemize}
\item \textbf{Mobile Bases:} Multiple locomotion platforms are supported, including the Segbot and the omnidirectional Ranger Mini.
\item \textbf{Robotic Arms:} The library includes over ten industrial and collaborative arms, such as the Realman RML, Franka Emika Panda, Universal Robots UR5, Flexiv Rizon, UFactory xArm, and the Elephant Robotics myCobot series.
\item \textbf{End Effectors:} A variety of interchangeable tools, including vacuum grippers and parallel-jaw grippers (e.g., Robotiq 2-Finger Adaptive Gripper 85 and DH Robotics AG95), are provided with standardized mounting interfaces.
\end{itemize}

\subsection{Simulation-Guided task formalization and skill learning}
Leveraging the generated simulation environments, we introduce a simulation-guided module for task formalization and skill acquisition. We first establish the THMM Benchmark, which decomposes language-conditioned instructions into structured sub-task sequences following the Sense–Plan–Act paradigm. Central to this framework is a modular skill architecture that encapsulates diverse robotic capabilities into reusable, hardware-agnostic primitives, enabling the system to handle both basic mobility and complex articulation manipulation. To support robust execution, a multimodal perception layer provides semantic scene understanding through the flexible integration of various vision backbones. Furthermore, a hybrid planning strategy combines symbolic task reasoning with continuous motion planning to ensure reliable operation in cluttered environments. Finally, a hierarchical management architecture orchestrates these components, ensuring scalable and conflict-free integration for complex, long-horizon mobile manipulation tasks.

\begin{figure}[t]
\centering
\includegraphics[width=\columnwidth]{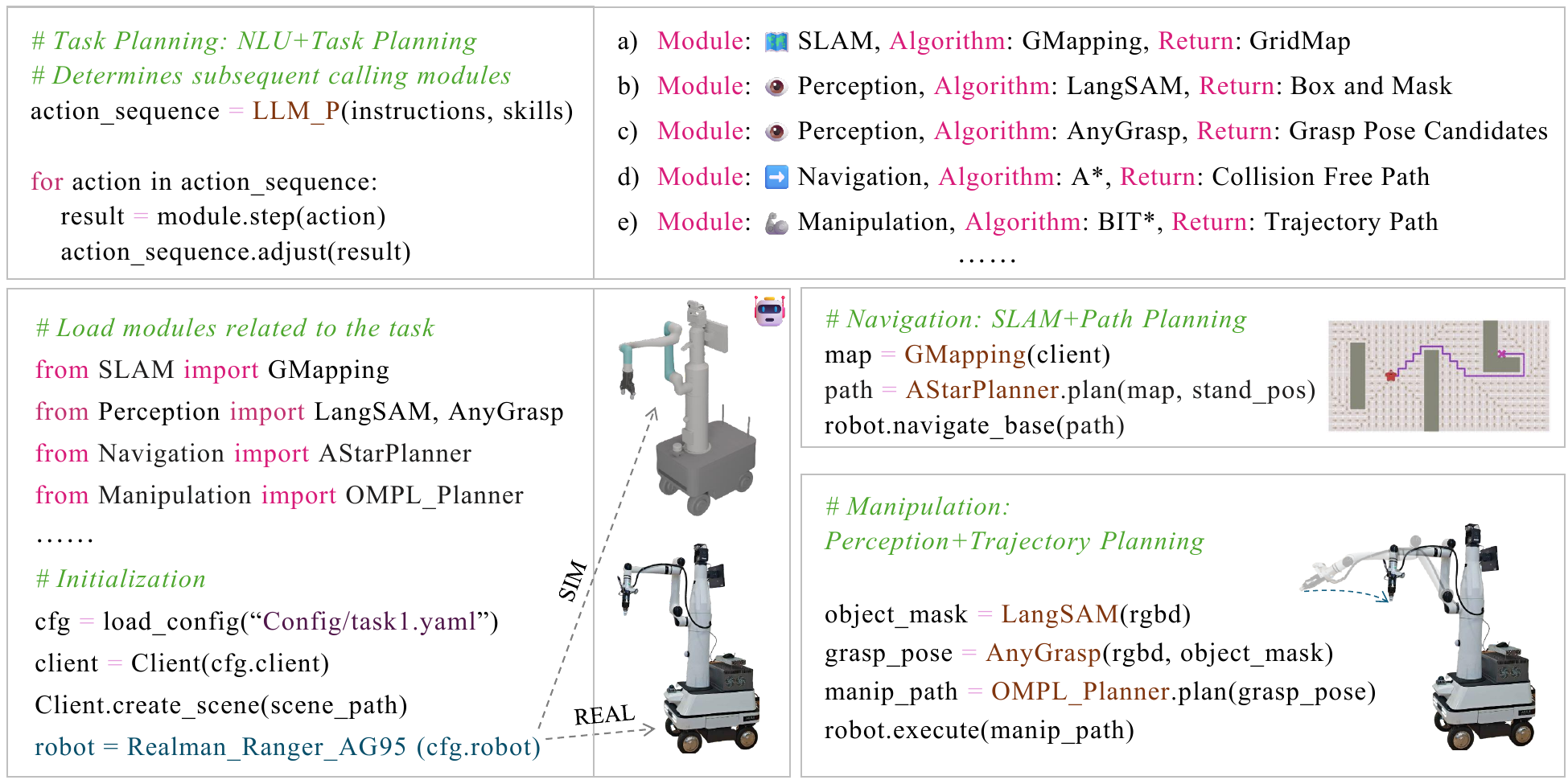}
\caption{Overview of the THMM task execution pipeline. The workflow illustrates the transition from high-level natural language instructions to low-level robotic execution, encompassing semantic task decomposition, scene-aware planning, and real-time trajectory generation for mobile manipulation.}
\label{fig:workflow}
\end{figure}

\subsubsection{THMM benchmark and task formalization} Complex household mobile manipulation tasks are typically driven by natural language instructions from humans, and their long-horizon nature poses significant challenges in terms of formalization and systematic evaluation. To address this, we propose the Transferable Household Mobile Manipulation (THMM) benchmark, which adopts the Sense–Plan–Act paradigm to decompose language-guided tasks into sequential sub-tasks that involve several fundamental robotic skills, with particular emphasis on perception and planning. To further illustrate this process, we present several task cases in Figure~\ref{fig:workflow}. Given the diversity of the required skills, our modular skill architecture emerges as an effective solution for handling multi-layer integration. This paper focuses on a representative task format for experimental evaluation, with detailed descriptions provided in Section~\ref{sec:experimental_settings}.

\begin{figure}[t]
\centering
\includegraphics[width=\linewidth]{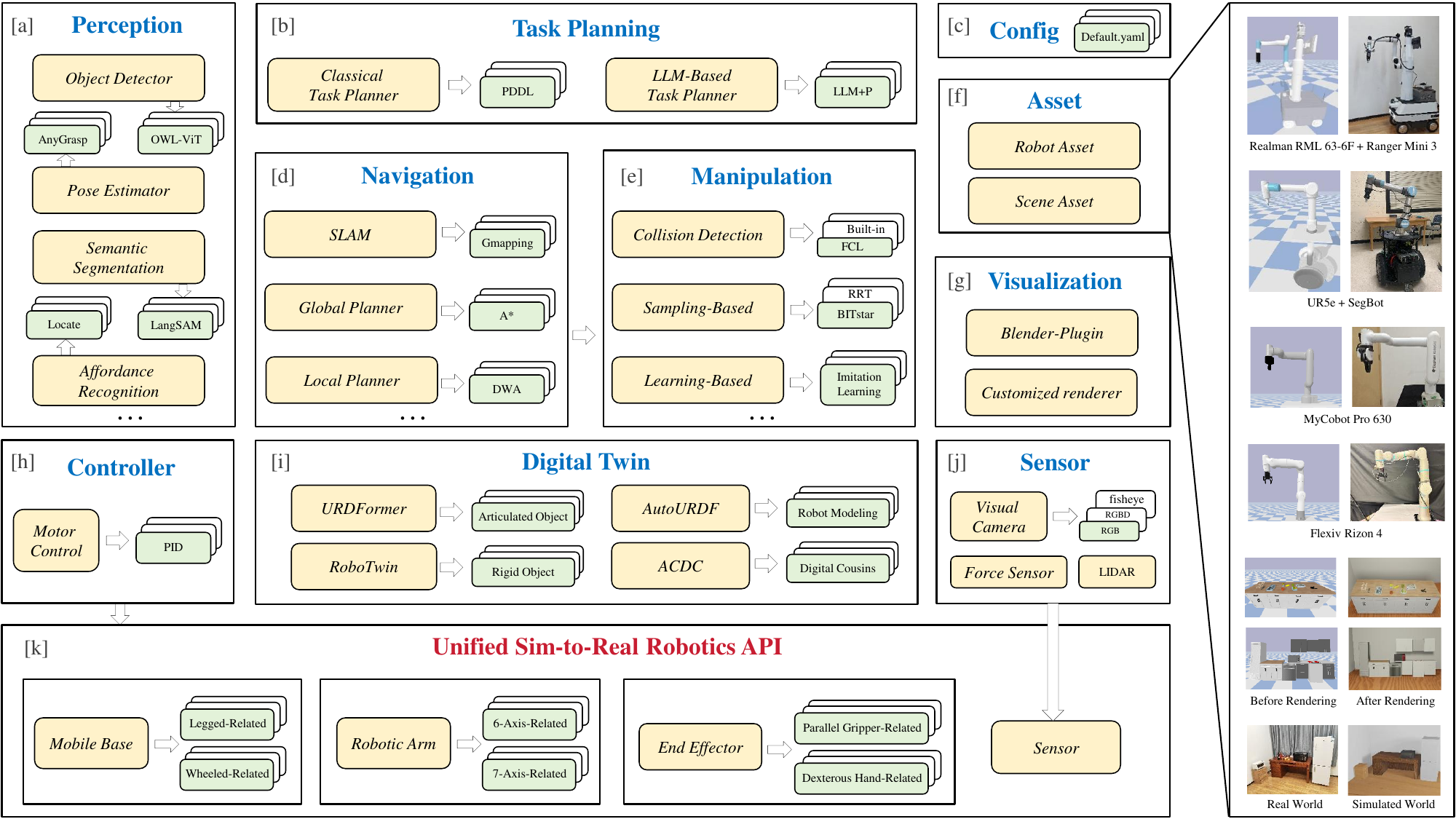}
\caption{Overview of \textbf{BestMan}’s modular skill architecture. The platform comprises eleven top-level modules (highlighted in blue and red): (a) Perception, (b) Task Planning, (c) Config, (d) Navigation, (e) Manipulation, (f) Asset, (g) Visualization, (h) Controller, (i) Digital Twin, (j) Sensor, and (k) Robotics API. Each module contains several submodules (highlighted in yellow), where various algorithms or contents (highlighted in green) can be implemented. The unified sim-to-real robotics API module is constructed based on the controller and sensor modules, while other modules are independent of these robotics API. The \textit{right panel} illustrates the platform’s applicability across various simulated and real-world mobile manipulators and environments.}
\label{fig:modular_skill_architecture}
\end{figure}

\subsubsection{Modular skill architecture}
Figure~\ref{fig:modular_skill_architecture} shows the modular skill architecture of \textbf{BestMan}, consisting of eleven top-level modules covering core functionalities such as perception, planning, control, and more. Each module contains submodules supporting different algorithms or functions. The unified sim-to-real robotics API connects the controller and sensor modules, enabling smooth transfer across various robotic platforms. This design promotes flexibility and scalability for both simulation and real-world mobile manipulation tasks.

Beyond these core modules, \textbf{BestMan} also includes valuable supplementary modules like visualization, control, collision detection, and SLAM, which enhance system usability and adaptability. To overcome rendering limitations of the Bullet engine, the platform integrates the Blender engine via dedicated plugins (Figure~\ref{fig:modular_skill_architecture}-g), significantly improving visual fidelity. The controller module (Figure~\ref{fig:modular_skill_architecture}-h) incorporates classic algorithms such as PID, MPC, and adaptive controllers for precise and stable motion control. Other supplementary modules likewise provide standard algorithms and extensible templates, allowing users to easily customize or extend functionalities for task-specific needs.

\begin{figure}[h]
\centering
\includegraphics[width=\linewidth]{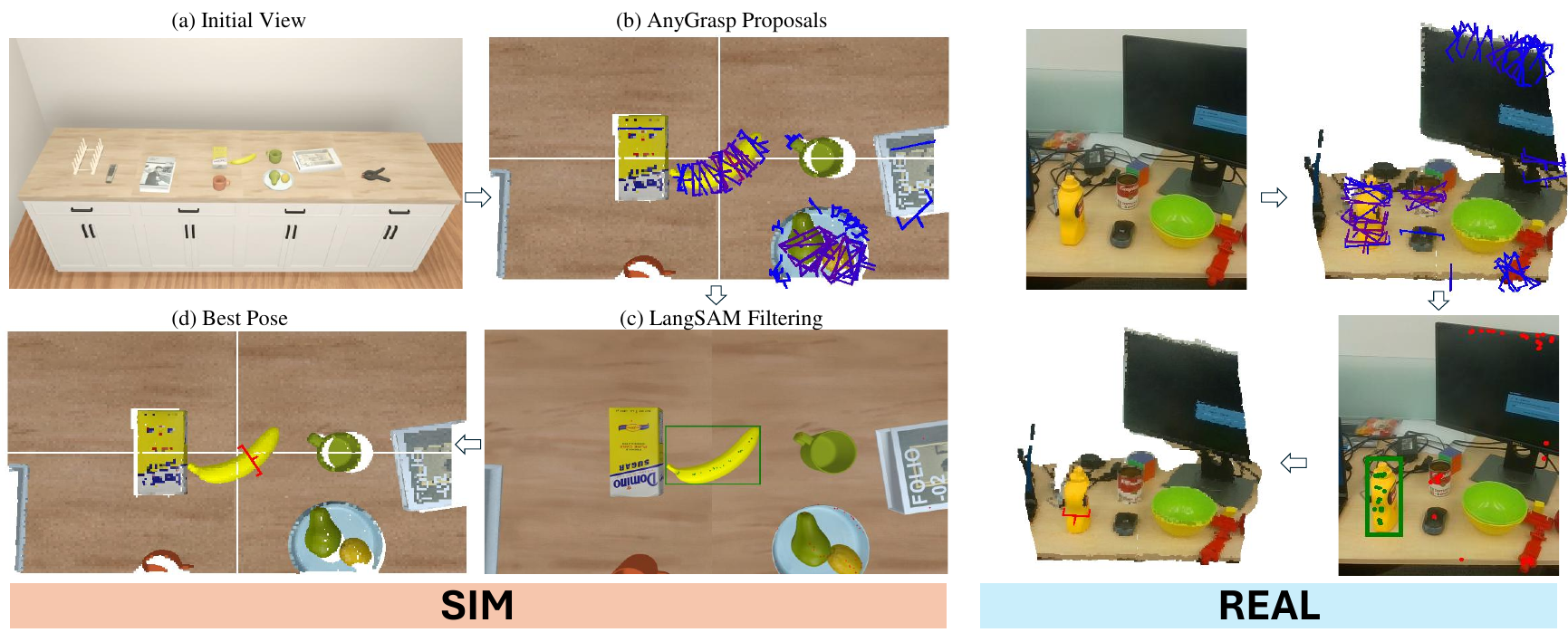}
\caption{Examples of combining LangSAM and AnyGrasp
algorithms to obtain optimal grasp pose from simulated and
real-world RGBD images.}
\label{fig:perception}
\end{figure}

\subsubsection{Multimodal perception for skill execution} In response to the cross-modal environmental perception challenge posed by sub-tasks, we design a multimodal perception layer that supports task-oriented sensory processing and semantic understanding, facilitating the flexible integration of state-of-the-art perception backbones. This layer connects perceptual inputs with the decision-making process, realizing the mapping \( P:\mathcal{S} \times \mathcal{M} \rightarrow \mathcal{O} \), where \(\mathcal{S}\) represents the observation space, \(\mathcal{M}\) denotes the semantic reasoning methods, and \(\mathcal{O}\) refers to the structured environmental representations. As shown in Figure~\ref{fig:perception}, the perception module facilitates a robust grasping pipeline that processes environment data through several key stages. This workflow demonstrates how BestMan maintains consistency between simulation and real world domains, allowing for seamless transition and high-precision execution in physical environments.
Each perception sub-module targets a specific aspect of environmental understanding: the object segmentation module utilizes LangSAM~\cite{langsam} and Grounded-SAM~\cite{grounded_sam} to obtain masks and bounding boxes; the grasp pose estimation module integrates AnyGrasp~\cite{anygrasp} and Contact-GraspNet~\cite{contact-graspnet} to predict actionable grasp poses. The affordance detection module employs Locate~\cite{locate} to identify the functional properties, among others. Ensembling all these sub-modules forms a coherent perception system that provides essential semantic support for the downstream planning and action process.

\subsubsection{Hybrid planning in cluttered environments} Motivated by the challenge of robust planning in cluttered environments, we implement a hybrid planning manner that integrates task and motion planning, providing a robust solution for both structured and unstructured environments.

For task planning, we integrate symbolic reasoning with large language models (LLMs), enabling effective generation of sub-tasks, even under ambiguous instructions, by leveraging the LLM's ability to handle complex, context-dependent tasks. We formalize this method into the following mathematical formulation:

\begin{equation}
\mathcal{P} = 
\underbrace{\mathit{\Gamma}_{\mathrm{symb}}(S)}_{\text{Symbolic Planner}} 
\circ 
\underbrace{\mathit{\Phi}_{\mathrm{LLM}}(I)}_{\text{LLM Interpreter`}},
\label{eq:hybrid_plan}
\end{equation}
where $\mathit{\Gamma}_{\text{symb}}$ is the symbolic planner, which
generates constraint-satisfaction problem solutions, $S = \{s_i\}_{i=1}^n$ is the PDDL-defined state space,
$\mathit{\Phi}_{\text{LLM}}$ is the LLM-based semantic parser, and $I$ is the natural language instruction. The operator $\circ$ is defined as:
\begin{equation}
x\circ y = 
\begin{cases} 
x \oplus y & \text{if } \mathcal{V}(x,y) \geq \theta , \\
\textsc{Replan}(feedback) & \text{otherwise},
\end{cases}
\label{eq:fusion}
\end{equation}
where $\oplus$ is the consensus-driven fusion process, $\mathcal{V}$ is the function that evaluates plan-semantic consistency and constraint satisfaction, and $\theta$ is the predefined threshold.

Then, we divide motion planning into navigation and manipulation, leveraging both classical and learning-based methods to meet task and environment complexity:


(1) For navigation, we use global path planning algorithms such as A*, RRT, and PRM to find optimal paths in known environments.
For dynamic or partially observable spaces, we use local methods like DWA and model predictive control (MPC) for real-time trajectory refinement.  

(2) For manipulation, we combine sampling-based methods from OMPL~\cite{ompl} (e.g., RRT, RRT*, and BIT*)  with reinforcement learning and imitation learning methods to enable adaptive grasping and object interaction across varied scene configurations.

\begin{figure}[t]
    \centering
    \includegraphics[width=\textwidth]{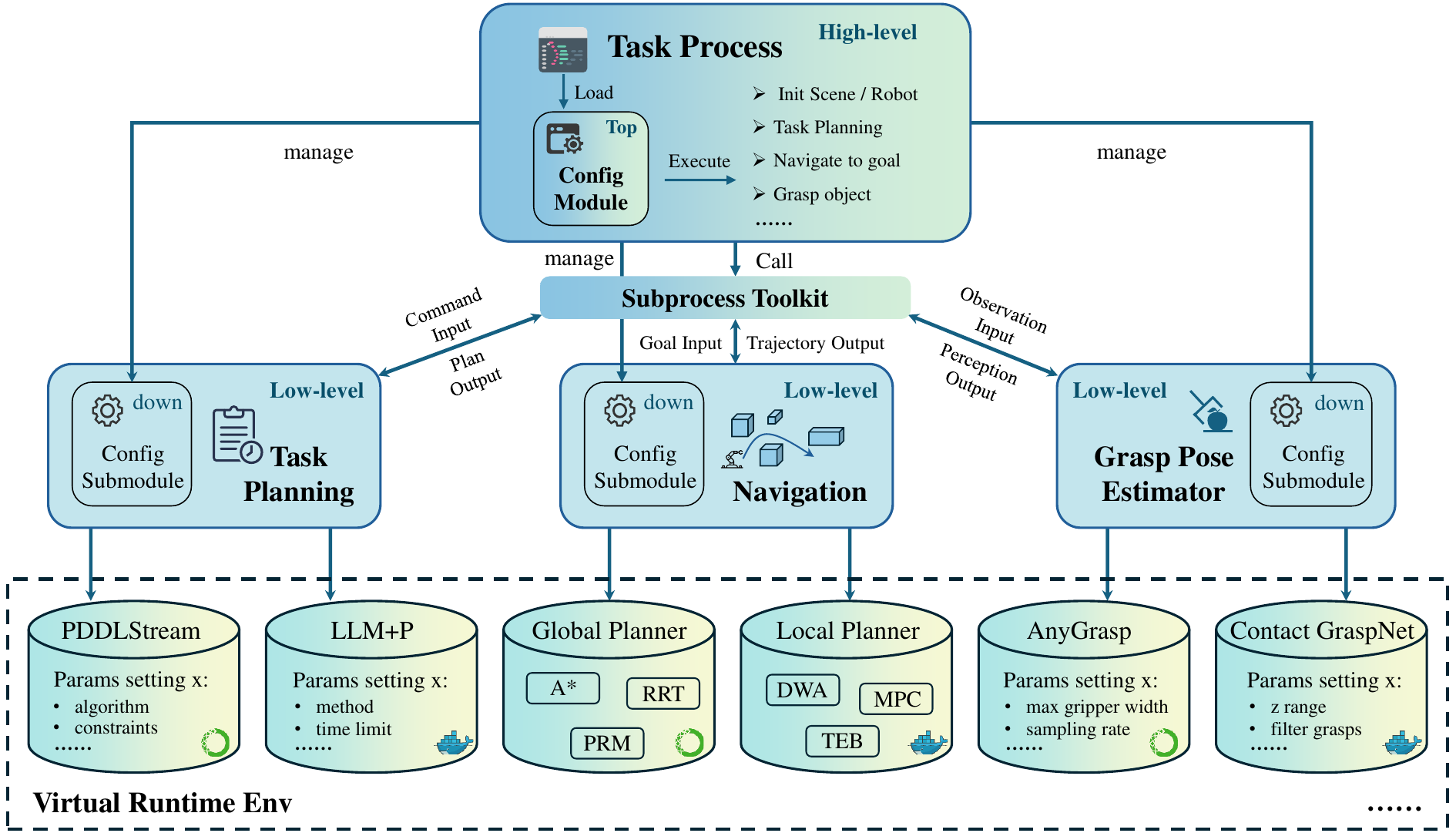}
    \caption{The hierarchical architecture for configuration and dependency management: 1) the top-level configuration module adopts a hierarchical structure to centrally manage the lower-level sub-modules, 2) the developed sub-process toolkit leverages containerization and virtual-environment technologies to isolate runtime dependencies.}
    \label{fig:config_dependency_management}
\end{figure}

\subsubsection{Hierarchical management and modular integration}
\label{sec:hierarchical_management}
As the platform scales, modular integration presents a critical challenge in managing configuration complexity and dependency conflicts, thereby threatening platform stability. To address this, we introduce a hierarchical architecture for configuration and dependency management, as illustrated in Figure~\ref{fig:config_dependency_management}.
Each functional module is paired with a dedicated configuration sub-module for algorithm-specific parameter tuning.
A global configuration module coordinates high-level settings across modules, providing a centralized interface for inspection, modification, and synchronization, thus reducing manual configuration errors.
Subsequently, to resolve dependency conflicts, 
we adopt a hybrid strategy combining containerization and virtual environments. Each module is encapsulated within an isolated runtime with on-demand execution, while the hierarchical communication mechanism ensures reliable data exchange. The hierarchical architecture facilitates scalable and conflict-free integration for our modular platform.

\subsection{HUM: Sim-to-Real deployments via unified middleware}
\label{sec:sim-to-real}

\begin{figure}[t]
    \centering
    \includegraphics[width=\textwidth]{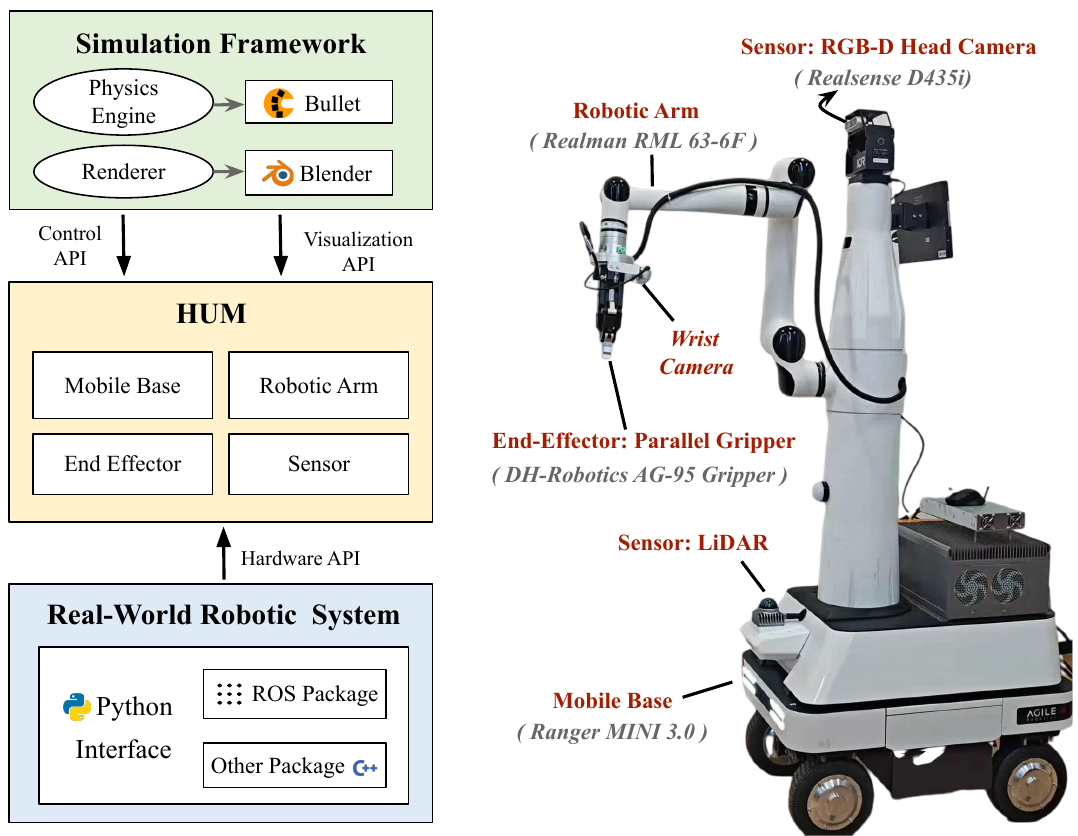}
    \caption{The HUM for sim-to-real deployments. It consists of two main parts: the unified APIs for mobile manipulators and the cross-platform compatibility and deployments. Each hardware component is sourced from a different manufacturer and features its own distinct interface library.}
    \label{fig:real_robot}
\end{figure}

As shown in Figure~\ref{fig:real_robot}, we introduce the HUM for seamless sim-to-real deployments, focusing on the mobile manipulator API design and the compatibility with heterogeneous hardware. 

\subsubsection{Unified api for mobile manipulators}
To overcome the interface discrepancies and enhance transfer robustness between the simulation framework and the real-world robotic system, we introduce HUM which provides an abstracted, aligned Python API architecture, as shown in Figure~\ref{fig:real_robot}. This middleware establishes a bijective mapping between the simulated and real-world domains, formalized as: $\mathcal{C}_{\text{sim}} \leftrightarrow \mathcal{C}_{\text{real}},\ \text{where}\ \mathcal{C} = \{ c_i \mid c_i \in \text{SE}(3) \times \mathbb{R}^n \}$, which enables both indirect algorithmic invocation and direct control over the mobile base, robotic arm, end effector, and sensor, without concern for the underlying implementation interfaces. It maintains consistent high-level commands, whether for simple base movements like \texttt{move\_forward()} or complex robotic arm control, such as \texttt{move\_eef\_to\_goal\_pose()}. Our design ensures seamless sim-to-real transfer without extensive modifications to accommodate ROS-based hardware-specific drivers, communication protocols, and safety mechanisms. By investing minimal effort to align with the middleware, extensive efforts are avoided to adapt the underlying system. 

\subsubsection{Cross-Platform Compatibility and Deployments} Another challenge lies in the limited compatibility with heterogeneous robotic hardware, caused by the tight coupling between real-world robotic systems (e.g., ROS) and specific simulation frameworks. To address this, our proposed HUM exhibits strong decoupling, making the platform both simulator- and hardware-agnostic. Such capability enables real-to-real transfer, allowing researchers to quickly deploy algorithms to their hardware without needing to set up identical hardware again, significantly improving the real-world reproducibility of mobile manipulation algorithms. What is more, it facilitates the exploration of algorithm robustness across various hardware setups.

\section{Experiments and analysis}
Our experiments aim to answer the following questions:
\begin{itemize}
\item{\textbf{Q1:}} Can the real-to-sim ASG module accurately generate high-quality simulated scenes from input images?
\item{\textbf{Q2:}} Is the modular design more effective than non-modular methods for complex tasks and environments?
\item{\textbf{Q3:}} Can HUM reduce errors and improve efficiency in sim-to-real deployments, and enhance compatibility of transferring to heterogeneous hardware?
\item{\textbf{Q4:}} Can \textbf{BestMan} efficiently complete mobile manipulation tasks based on the real-to-sim-to-real pipeline?
\end{itemize}

\begin{sidewaystable}
\setlength{\tabcolsep}{2pt}
\caption{Qualitative comparison among \textbf{BestMan} and other robotic platforms, grouped into four categories from top to bottom: simulators, simulation frameworks, real-world platforms, and our proposed \textbf{BestMan}. Compared to several platforms, our \textbf{BestMan} offers versatile functions, enables automated scene generation, provides flexible configurations, supports multiple tasks, and provides better scalability and adaptability.}\label{tab:platform_comparision}
\begin{tabular}{lcccccccccccc} 
\toprule
\multirow{2}{*}{Platform} & 
\multirow{2}{*}{\begin{tabular}[c]{@{}c@{}}Scene\\Generation\footnotemark[1]\end{tabular}} & 
\multirow{2}{*}{\begin{tabular}[c]{@{}c@{}}Physics\\Backends\footnotemark[2]\end{tabular}} &
\multirow{2}{*}{\begin{tabular}[c]{@{}c@{}}Hardware\\ Reconfiguration\footnotemark[3]\end{tabular}} &
\multicolumn{6}{c}{Tasks\footnotemark[4]} & 
\multirow{2}{*}{\begin{tabular}[c]{@{}c@{}}Modular\\Scalability\footnotemark[7]\end{tabular}} & 
\multirow{2}{*}{\begin{tabular}[c]{@{}c@{}}Sim2Real API\\Consistency\end{tabular}} &
\multirow{2}{*}{\begin{tabular}[c]{@{}c@{}}Hardware\\Adaptability\footnotemark[8]\end{tabular}} \\
\cmidrule(lr){5-10}
 &  &  &  & V. & Nav. & Manip.\footnotemark[5] & MRC & Long-H. & Cust.\footnotemark[6] &  &  &  \\  
\midrule
Gazebo~\cite{gazebo} & \bloodredxmark & O, B, S, D & \bloodredxmark & \bloodredxmark & \bloodredxmark & Obj & \bloodredxmark & \bloodredxmark & \bloodredxmark & \bloodredxmark & \bloodredxmark & \bloodredxmark \\
PyBullet~\cite{pybullet} & \bloodredxmark & B & \bloodredxmark & \bloodredxmark & \bloodredxmark & Obj & \bloodredxmark & \bloodredxmark & \bloodredxmark & \bloodredxmark & \bloodredxmark & \bloodredxmark \\
MuJoCo~\cite{mujoco} & \bloodredxmark & Custom & \bloodredxmark & \bloodredxmark & \bloodredxmark & Obj & \bloodredxmark & \bloodredxmark & \bloodredxmark & \bloodredxmark & \bloodredxmark & \bloodredxmark \\
SAPIEN~\cite{sapien} & \cautionyellowcheckodd & P & \bloodredxmark & \bloodredxmark & \bloodredxmark & Part & \bloodredxmark & \bloodredxmark & \bloodredxmark & \bloodredxmark & \bloodredxmark & \bloodredxmark \\ 
V-REP~\cite{V-REP} & \bloodredxmark & O, B, V, N & \bloodredxmark & \bloodredxmark & \bloodredxmark & Obj & \bloodredxmark & \bloodredxmark & \bloodredxmark & \bloodredxmark & \bloodredxmark & \bloodredxmark \\ 
\midrule
AI2-THOR~\cite{AI2-THOR} & \cautionyellowcheckodd & Unity & \bloodredxmark & \OliveGreencheck & \OliveGreencheck & Obj & \OliveGreencheck & \bloodredxmark & \bloodredxmark & \bloodredxmark & \bloodredxmark & \bloodredxmark \\
VirtualHome~\cite{virtualhome} & \cautionyellowcheckodd & Unity & \bloodredxmark & \OliveGreencheck & \OliveGreencheck & Obj & \bloodredxmark & \bloodredxmark & \bloodredxmark & \bloodredxmark & \bloodredxmark & \bloodredxmark \\
RLBench~\cite{Rlbench} & \cautionyellowcheckodd & B & \bloodredxmark & \OliveGreencheck & \bloodredxmark & Obj & \bloodredxmark & \bloodredxmark & \OliveGreencheck & \bloodredxmark & \bloodredxmark & \bloodredxmark \\
 Habitat~\cite{habitat} & \cautionyellowcheckodd & B & \bloodredxmark & \OliveGreencheck & \OliveGreencheck & \bloodredxmark & \bloodredxmark & \OliveGreencheck & \bloodredxmark & \bloodredxmark & \bloodredxmark & \bloodredxmark \\
robosuite~\cite{robosuite} & \cautionyellowcheckodd & Mujoco & \cautionyellowcheckodd & \bloodredxmark & \bloodredxmark & Obj & \OliveGreencheck & \bloodredxmark & \OliveGreencheck & \bloodredxmark & \bloodredxmark & \bloodredxmark \\
iGibson~\cite{igibson} & \cautionyellowcheckodd & B & \bloodredxmark & \bloodredxmark & \OliveGreencheck & Obj & \OliveGreencheck & \OliveGreencheck & \OliveGreencheck & \bloodredxmark & \bloodredxmark & \bloodredxmark \\
Orbit~\cite{orbit} & \cautionyellowcheckodd & P & \cautionyellowcheckodd & \bloodredxmark & \OliveGreencheck & Part & \bloodredxmark & \OliveGreencheck & \OliveGreencheck & \bloodredxmark & \bloodredxmark & \bloodredxmark \\
ManiSkill~\cite{ManiSkill} & \cautionyellowcheckodd & SAPIEN & \bloodredxmark & \OliveGreencheck & \bloodredxmark & Part & \OliveGreencheck & \bloodredxmark & \bloodredxmark & \bloodredxmark & \bloodredxmark & \bloodredxmark \\
RoboCasa~\cite{robocasa} & \cautionyellowcheckodd & Mujoco & \cautionyellowcheckodd & \bloodredxmark & \bloodredxmark & Part & \bloodredxmark & \OliveGreencheck & \OliveGreencheck & \OliveGreencheck & \bloodredxmark & \bloodredxmark \\
Kitchen Worlds~\cite{kitchen_worlds} & \cautionyellowcheckodd & B & \bloodredxmark & \bloodredxmark & \OliveGreencheck & Part & \bloodredxmark & \OliveGreencheck & \bloodredxmark & \bloodredxmark & \bloodredxmark & \bloodredxmark \\
\midrule
HomeRobot~\cite{homerobot} & \cautionyellowcheckodd & Habitat & \bloodredxmark & \OliveGreencheck & \OliveGreencheck & Obj & \bloodredxmark & \bloodredxmark & \bloodredxmark & \OliveGreencheck & \OliveGreencheck & \bloodredxmark \\
OK-Robot~\cite{ok-robot} & \bloodredxmark & \bloodredxmark & \bloodredxmark & \OliveGreencheck & \OliveGreencheck & Obj & \bloodredxmark & \bloodredxmark & \bloodredxmark & \OliveGreencheck & \OliveGreencheck & \bloodredxmark \\
\midrule
\textbf{BestMan (ours)} & \OliveGreencheck & Dynamic & \OliveGreencheck & \OliveGreencheck & \OliveGreencheck & Part & \OliveGreencheck & \OliveGreencheck & \OliveGreencheck & \OliveGreencheck & \OliveGreencheck & \OliveGreencheck \\
\bottomrule
\end{tabular}

\footnotetext{\bloodredxmark \quad Not available \quad \cautionyellowcheckodd \quad Partial available \quad \OliveGreencheck \quad Fully available}

\footnotetext[1]{~\cautionyellowcheckodd means rule-based procedural scene generation, while \OliveGreencheck means automated real-to-sim scene generation.}

\footnotetext[2]{~\textbf{O} for \textbf{O}DE, \textbf{B} for \textbf{B}ullet, \textbf{S} for \textbf{S}imbody, \textbf{D} for \textbf{D}ART, \textbf{P} for \textbf{P}hysX, \textbf{V} for \textbf{V}ortex, \textbf{N} for \textbf{N}ewton.}

\footnotetext[3]{~\cautionyellowcheckodd means limited to predefined assemblies; \OliveGreencheck allows free hardware replacement and reconfiguration.}

\footnotetext[4]{~V.: Vision; Nav.: Navigation; Manip.: Manipulation; MRC: Multi-Robot Collaboration; Long-H.: Long-horizon; Cust.: Customizable.}

\footnotetext[5]{~Obj means only object-level interaction is supported. Part means part-level interaction is supported.}

\footnotetext[6]{~Tasks can be customized beyond predefined categories to suit specific user requirements.}

\footnotetext[7]{~Modular Scalability means rapidly integrating new algorithms or modules into the platform’s task pipelines.}

\footnotetext[8]{~Hardware Adaptability refers to the ability to transfer strategies across heterogeneous hardware.}
\end{sidewaystable}

\subsection{Experimental settings}
\label{sec:experimental_settings}

We formulate the THMM task as a language-guided instruction: \textit{``Move the object from the start\_receptacle to the goal\_receptacle"}. The object is a small household rigid item (e.g., a can of cola or a bowl), and the receptacle is a large articulated piece of furniture where the object can either
be placed on the surface or inside the container, with the robot positioned at a random initial pose. We decompose the THMM task into three key sub-tasks: (1) {\em Approach}, in which the robot detects and navigates towards the target object; (2) {\em Manipulation}, in which the robot interacts with the start receptacle and grasps the object; and (3) {\em Placement}, in which the robot transports the object to the goal receptacle and places it at the specific location. These sub-tasks require 6 fundamental skills, including \texttt{Find}, \texttt{Navigation}, \texttt{Grasp Estimation}, \texttt{Open}, \texttt{Pick}, and \texttt{Place}. We further define three levels of task difficulty: (1) \textbf{Easy}, the initial and target locations are both on the surface of receptacles; (2) \textbf{Medium}, the initial and target locations are inside receptacles that are already open; and (3) \textbf{Hard}, the initial and target locations are inside closed receptacles.

For the experimental scenes, building upon the outputs of the real-to-sim pipeline, we created more than 100 interactive simulated scenes by applying serveral augmentation techniques, including visual, physics, kinematic (scale and pose), and instance randomization. Additionally, real-world experiments were conducted in a single-room environment styled as a kitchen, as shown in Figure~\ref{fig:overview}, featuring a refrigerator, oven, cabinets, table, and other elements.

For real-world experiments, we employ multiple mobile manipulator setups. The primary setup, as shown in Figure\mbox{~\ref{fig:real_robot}}, consists of a Ranger MINI 3.0 mobile base, a Realman RML 63-6F robotic arm, and a DH-Robotics AG-95 parallel gripper. The sensory suite includes a LiDAR and two Intel RealSense D435i RGB-D cameras (wrist-mounted and head-mounted). Computational tasks are handled by an onboard workstation equipped with an NVIDIA RTX 3090 GPU to ensure high-performance real-time processing. Notably, while our primary experiments leverage high-end hardware, the platform's minimum recommended specification is an NVIDIA RTX 2080 Ti GPU, ensuring broad accessibility for various research deployment scenarios.

\subsection{Qualitative analysis}\label{qualitative analysis}
Table~\ref{tab:platform_comparision} provides a qualitative comparison between \textbf{BestMan} and several representative robotic frameworks. In contrast to existing systems, \textbf{BestMan} delivers comprehensive support for diverse environmental assets, various physics backends, and modular robot assembly. Beyond these technical features, our platform distinguishes itself through its capacity for automated scene generation and flexible configuration across a wide range of task categories. As illustrated in Table~\ref{tab:platform_comparision}, these attributes underscore the superior scalability and adaptability of \textbf{BestMan}, making it a robust foundation for advanced research in mobile manipulation.

\begin{table*}[t]
\caption{Accuracy analysis of real-to-sim ASG. 
We list three scenes transferred using ASG: bedroom, living room, and kitchen. The metrics evaluate object structure accuracy (Category, Modeling, Joint), pose errors (L2 Distance, Orientation Difference), and scale errors (Bounding Box IOU, Center IoU). C. Score is a composite score based on human evaluation.}\label{tab:sim-to-sim and real-to-sim reconstruction}
\resizebox{\textwidth}{!}{
\begin{tabular}{@{}l|ccccccc|ccc@{}}
\toprule
\multirow{2}[1]{*}{Scene Type} & \multicolumn{7}{c|}{\textbf{Simulation Image Input}} & \multicolumn{3}{c}{\textbf{Real-World Image Input}} \\
\cmidrule(lr){2-8} \cmidrule(lr){9-11}
 & Cat. $\uparrow$ & Mod. $\uparrow$ & Joint. $\uparrow$ & L2 Dist. (cm) $\downarrow$ & Ori. Diff. (rad) $\downarrow$ & Bbox IoU $\uparrow$ & Cen. IoU $\uparrow$ & Cat. $\uparrow$ & Mod. $\uparrow$ & C. Score $\uparrow$ \\ 
\midrule
Bedroom & 100\% & 94\% & 90\% & $5.20 \pm 2.85$ & $0.07 \pm 0.06$ & $0.68 \pm 0.19$ & $0.75 \pm 0.15$ & 100\% & 91\% & 90\% \\ 
\midrule
Living Room & 100\% & 93\% & 91\% & $6.45 \pm 3.90$ & $0.06 \pm 0.05$ & $0.71 \pm 0.18$ & $0.76 \pm 0.14$ & 100\% & 88\% & 85\% \\
\midrule
Kitchen & 100\% & 92\% & 88\% & $4.85 \pm 2.60$ & $0.05 \pm 0.04$ & $0.72 \pm 0.20$ & $0.78 \pm 0.16$ & 100\% & 86\% & 89\% \\
\botrule
\end{tabular}}
\end{table*}

\begin{table}[t]
    \caption{Comparison of articulated scene generation methods. Input images are sourced from both simulation and the real world. The metrics evaluate object joint accuracy, pose error,(combining L2 distance and orientation difference), and center IoU. ASG outperforms URDFormer and ACDC in both inputs, with more accurate and stable generation.}\label{tab:reconstruction comparision}
    \begin{tabular}{@{}l|c|ccc@{}}
    \toprule
    Method & Input & \textbf{Joint.} $\uparrow$ & \textbf{Pose Err.} $\downarrow$ & \textbf{Cen. IOU} $\uparrow$ \\
    \midrule
    \multirow{2}[1]{*}{URDFormer} & Sim & 88\% & $7.65 \pm 3.18$ & $0.65 \pm 0.18$ \\
      & Real & 81\% & $9.24 \pm 4.55 $ & $0.58 \pm 0.14$ \\
    \midrule
    \multirow{2}[1]{*}{ACDC} & Sim & 78\% & $5.21 \pm 2.16$ & $0.71 \pm 0.15$ \\
      & Real & 75\% & $6.13 \pm 2.78$ & $0.68 \pm 0.12$ \\
    \midrule
    \multirow{2}[1]{*}{\textbf{ASG (Ours)}} & Sim & \textbf{90\%} & \bm{$3.11 \pm 1.54$} & \bm{$0.80 \pm 0.06$} \\
      & Real & \textbf{85\%} & \bm{$4.02 \pm 1.40$} & \bm{$0.72 \pm 0.04$} \\
    \botrule
    \end{tabular}
\end{table}

\subsection{Accuracy analysis of ASG}
\label{sec:real-to-sim analysis}
To rigorously assess the performance of ASG module in sim-to-sim scenarios, we employ a comprehensive suite of quantitative metrics. Geometric and semantic fidelity are measured by Categorization (Cat.) and Modeling (Mod.) Accuracy, which evaluate object identification and reconstruction quality, respectively. Spatial precision is quantified via L2 Distance, Orientation Difference, and 3D Bounding Box Intersection over Union (Bbox IoU), alongside Center-aligned IoU (Cen. IoU) to isolate shape overlap from localization errors. Furthermore, Joint Accuracy (Joint.) evaluates the structural integrity of articulated components, including joint types, motion ranges, and connectivity. Finally, a composite Pose Error, defined as $E_{\text{pose}} = \alpha \cdot d + \beta \cdot \theta$, provides a unified assessment by weighting translation $d$ and rotation $\theta$ through coefficients $\alpha$ and $\beta$.For real-to-sim evaluation, we conducted a human study involving five domain experts who performed blind assessments of reconstructed scenes. Evaluators assigned scores (0–100) based on three standardized criteria: object recognition accuracy, geometric completeness, and joint structure fidelity.

To answer \textbf{Q1}, we analyze the accuracy of the ASG module. Table~\ref{tab:sim-to-sim and real-to-sim reconstruction} shows that \textbf{BestMan} achieves superior scene reconstruction performance across three types. For simulation inputs, it attains perfect category accuracy (100\%), modeling accuracy (up to 94\%), joint accuracy of 90\%, and precise geometric alignment, with lower errors, bounding box IoU, and center IoU. In more challenging real-world inputs, we further report a rubric-based human evaluation (denoted as C. Score). Unlike simulation inputs where ground-truth 3D scene geometry and articulated structures are fully accessible for quantitative metrics, real-world images do not provide such complete 3D annotations. Therefore, C. Score is used to capture semantic fidelity and perceptual plausibility from the perspective of human raters. Human evaluation confirms high semantic fidelity, with category accuracy remaining at 100\% and modeling accuracy above 85\%, and a composite score exceeding 89\%. 

As shown in Table~\ref{tab:reconstruction comparision}, comparative results demonstrate that our module significantly outperforms baselines such as URDFormer~\cite{urdformer} and ACDC~\cite{acdc}, particularly in joint modeling and pose estimation. This superiority is largely attributed to our hierarchical structural parsing approach and refined modeling of articulated parameters, which enhance stability in complex scenes. However, error analysis reveals that challenges persist in scenarios involving multi-layer occlusions and texture homogeneity. Future work will aim to mitigate these issues by integrating multi-view fusion and incorporating richer prior knowledge to further improve reconstruction robustness.

\begin{figure}[t]
\centering
\includegraphics[width=\columnwidth]{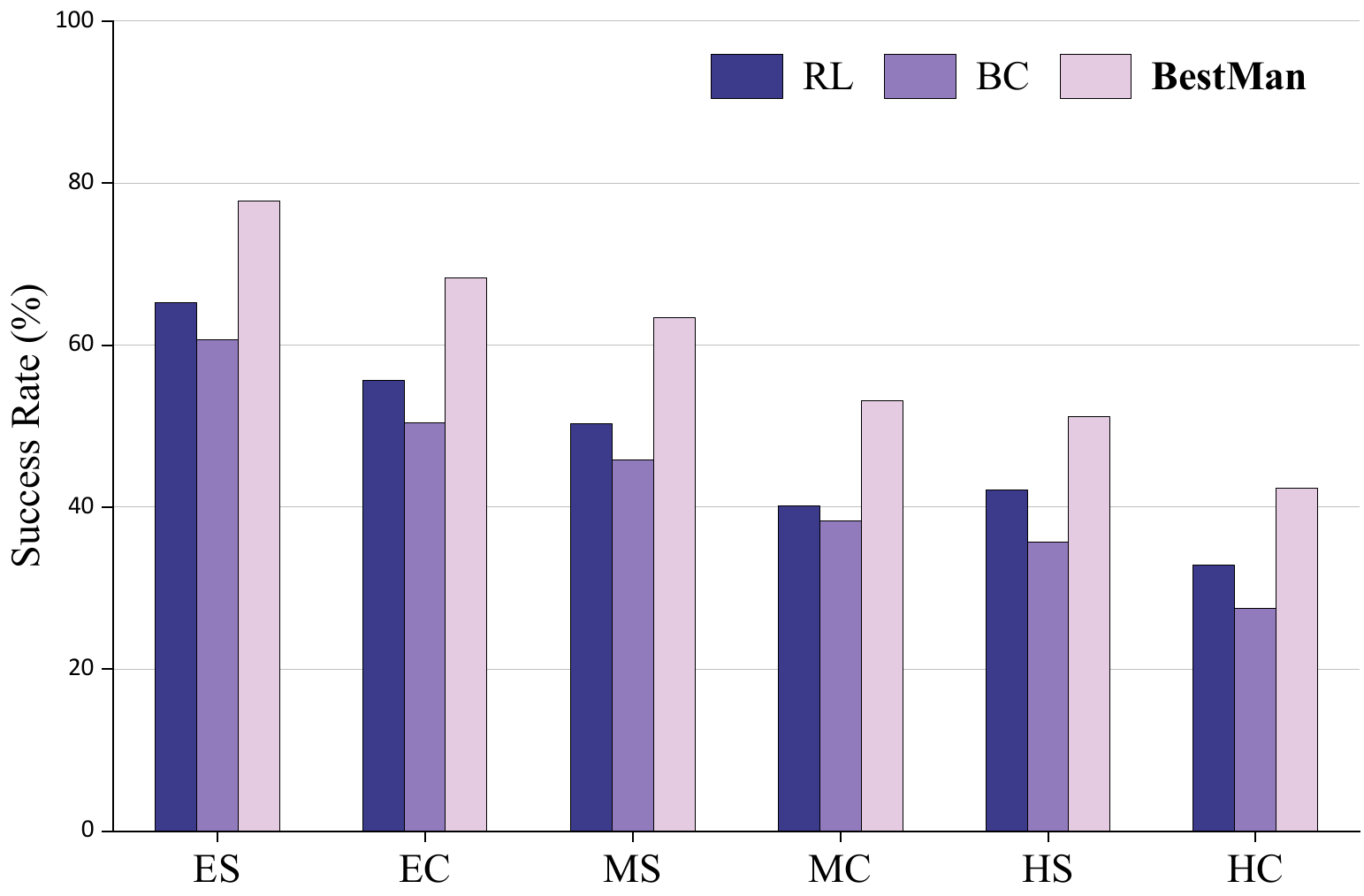}
\caption{
Comparison between \textbf{BestMan}'s modular architecture and non-modular baselines, including Reinforcement Learning (RL) and Behavior Cloning (BC).
The evaluation covers three tasks in two environments. E/M/H denote task difficulty levels (Easy, Medium, Hard), and S/C represent environment complexity (Simple, Complex).
}\label{fig:modular_comparison}
\end{figure}

\subsection{Effectiveness of modular design}
\label{sec:modular design evaluation}
To answer \textbf{Q2}, we evaluate the effectiveness of our modular design across six settings in ASG-generated simulated scenes, with 10 trials per scene type. As shown in Figure~\ref{fig:modular_comparison}, it consistently outperforms Reinforcement Learning and Behavior Cloning baselines across all settings. The performance gap increases with task difficulty and environmental complexity; notably, in the hardest setting (HC), our method achieves a 42.3\% success rate, compared to 32.8\% for RL and 27.5\% for BC. These results demonstrate the superiority of modular design for complex tasks and environments.

\begin{table}[t]
\centering
\caption{Efficiency comparison between our HUM middleware and ROS for sim-to-real transfer on two robotic hardware setups (RH1, RH2), evaluated by transfer time (TT), response time (RT), and task success rate (SR).}\label{tab:sim-to-real study}
\begin{tabular}{l|cccccc}
\toprule
\multirow{2}{*}{Framework} & \multicolumn{2}{c}{\textbf{TT (min)} $\downarrow$} & \multicolumn{2}{c}{\textbf{RT (ms)} $\downarrow$} & \multicolumn{2}{c}{\textbf{SR (\%)} $\uparrow$} \\
\cmidrule(lr){2-3} \cmidrule(lr){4-5} \cmidrule(lr){6-7}
& RH1 & RH2 & RH1 & RH2 & RH1 & RH2 \\
\midrule
ROS & 65 & 78 & 5.8 & 6.2 & 47.5 & 47.8 \\
\textbf{HUM (Ours)} & \textbf{14} & \textbf{18} & \textbf{4.5} & \textbf{5.3} & 45.2 & \textbf{49.0} \\
\botrule
\end{tabular}
\end{table}

\subsection{Efficiency comparison of HUM}
\label{sec:sim-to-real study}
To answer \textbf{Q3}, this section evaluates the efficiency and compatibility of our HUM for sim-to-real deployments across heterogeneous robotic hardware. Specifically, the transfer time (TT) encompasses the cumulative duration required for environment configuration, dependency resolution, driver compilation, and node initialization. As shown in Table~\ref{tab:sim-to-real study}, our HUM significantly reduces sim-to-real deployment time, achieving a 3$\times$ to 5$\times$ improvement over ROS (14 min vs. 65 min on RH1, and 18 min vs. 78 min on RH2). It also consistently exhibits lower response latency, with reductions of 1.3 ms and 0.9 ms on RH1 and RH2, respectively, which can be attributed to its lightweight nature. While the success rate remains comparable to ROS on RH1, our HUM slightly outperforms ROS on RH2 (49.0\% vs. 47.8\%). These results indicate that the proposed HUM module enhances the efficiency of the sim-to-real deployments while maintaining competitive task performance in real-world deployments.

\begin{table*}[!t]
\centering
\caption{
Results on the THMM Benchmark in simulation and real-world settings, where \textbf{BestMan} executes three sub-tasks, including Approach, manipulation, and placement. Six skills are grouped into four categories, each using a strategy type: H (heuristic), F (fine-tuned), or R (reinforcement learning).  Metrics include sub-task success rate (SR), average sub-task success rate (ASR), overall success rate (OSR), and average time steps (ATS) to complete each task.
}\label{tab:THMM Benchmark}
\resizebox{\textwidth}{!}{
\begin{tabular}{cccccccc}
\toprule
\multirow{2}[1]{*}{Environment} & \multirow{2}[1]{*}{Skill Strategy} & \multicolumn{3}{c}{\textbf{Sub-task SR} (\%) $\uparrow$} & \multirow{2}[1]{*}{\textbf{ASR} (\%) $\uparrow$} & \multirow{2}[1]{*}{\textbf{OSR} (\%) $\uparrow$ } & \multirow{2}[1]{*}{\textbf{ATS} $\downarrow$ } \\
\cmidrule{3-5}
  &  & Approach & Manipulation & Placement \\
\midrule
\multirow{5}[1]{*}{\textit{Simulation}} 
& \makebox[2cm][s]{H - H - H - H} & 71.2 & 58.3 & 48.3 & 59.3 & 38.5 & 186 \\
& \makebox[2cm][s]{F - H - F - H} & 80.5 & 60.1 & 53.9 & 64.8 & 45.6 & \textbf{179} \\
& \makebox[2cm][s]{H - R - H - R} & 83.3 & 72.2 & 56.8 & 70.8 & 50.3 & 195 \\
& \makebox[2cm][s]{F - R - F - R} & 88.7 & 75.5 & 65.2 & 76.5 & 60.7 & 189 \\
& \makebox[2cm][s]{F - R - R - R} & 87.4 & 80.9 & 72.4 & \textbf{80.2} & \textbf{66.2} & 201 \\
\midrule
\multirow{5}[1]{*}{\textit{Real World}} 
& \makebox[2cm][s]{H - H - H - H} & 67.4 & 51.2 & 30.7 & 49.8 & 23.4 & 209 \\
& \makebox[2cm][s]{F - H - F - H} & 75.6 & 55.9 & 38.5 & 56.7 & 30.3 & \textbf{204} \\
& \makebox[2cm][s]{H - R - H - R} & 70.2 & 65.7 & 48.1 & 61.3 & 42.5 & 215 \\
& \makebox[2cm][s]{F - R - F - R} & 80.1 & 72.2 & 57.3 & \textbf{69.9} & \textbf{52.2} & 210 \\
& \makebox[2cm][s]{F - R - R - R} & 80.7 & 63.2 & 45.3 & 63.1 & 39.8 & 226 \\
\botrule
\end{tabular}}
\end{table*}

\begin{figure}[t!]
    \centering
    \includegraphics[width=\linewidth]{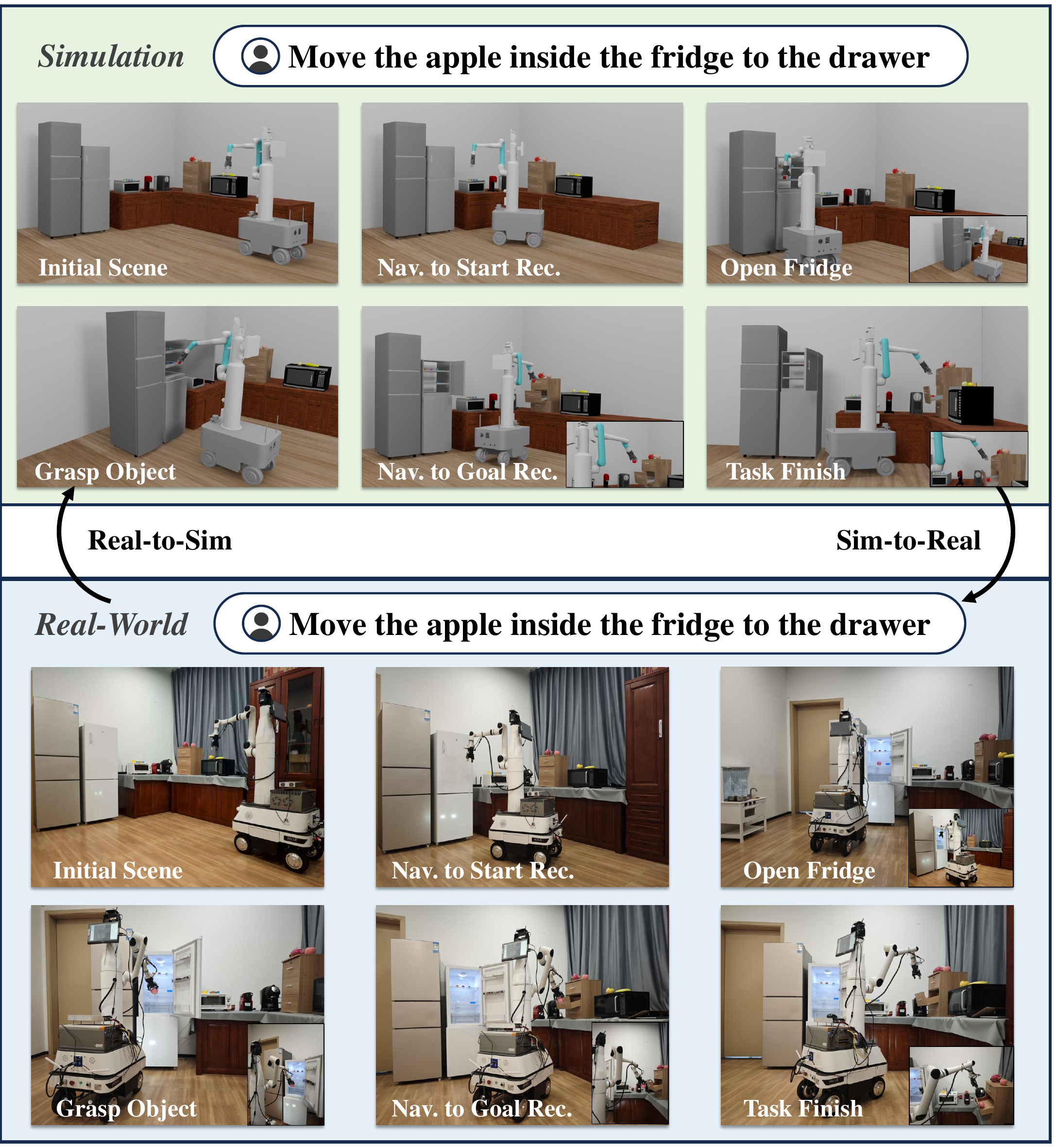}
    \caption{
    Real-to-sim-to-real pipeline of the \textbf{BestMan} platform. Real-world environments are first mapped into a high-fidelity simulation using the ASG module. 
    Once the skills are learned, they are seamlessly deployed back to the physical robot operating in the original real-world environment.
    }
    \label{fig:user cases}
\end{figure}

\subsection{Overall verification of BestMan on real-world tasks}
\label{sec:overall_verification}


To answer \textbf{Q4}, this section presents quantitative results of different skill strategies on the THMM task. The heuristic baseline uses off-the-shelf models and heuristic policies, the fine-tuned baseline adapts pre-trained models with collected simulated data, and the RL baseline learns policies through direct interaction in simulation.
Table~\ref{tab:THMM Benchmark} indicates that hybrid strategies integrating prior knowledge through fine-tuning with adaptive policy learning via reinforcement (e.g., F-R-F-R) consistently outperform purely heuristic method and exhibit greater reliability than RL-only strategies. These findings suggest that the \textbf{BestMan} platform leverages complementary strengths, enabling scalable, robust deployments and systematic refinement in complex real-world tasks.



Figure~\ref{fig:user cases} illustrates the complete real-to-sim-to-real cycle enabled by the \textbf{BestMan} platform. As shown in the upper part of the figure, it first seamlessly maps the real-world environments into a high-fidelity simulated scene using the ASG module, which reconstructs accurate geometry, articulated structures, and physical parameters. Within this reconstructed scene, different hybrid strategies integrating multiple skills for mobile manipulation can be rapidly composed, evaluated, and optimized across diverse tasks and environments. Once the strategies are learned, as shown in the bottom part, they are seamlessly deployed back to the physical robot, achieving superior performance in the original real-world environment. This bidirectional mapping enables the mobile manipulator to execute language-guided tasks effectively, demonstrating \textbf{BestMan}’s capability for efficient skill learning and robust real-world generalization.

\section{Application}\label{appendix_secC}
Building on \textbf{BestMan}'s strengths, researchers can explore and apply it in the following areas:

\begin{itemize}
\item \textbf{Visual Navigation}: The \textbf{BestMan}'s sensor, perception, and control module supports the mobile manipulator in acquiring environmental information and navigating in real time in complex scenarios.
\item \textbf{Long-horizon Tasks}: Tasks such as cleaning, item sorting, and handling are broken down into a series of subtasks by \textbf{BestMan}'s efficient task planning module, which ensures task consistency at all stages and executes them through dedicated submodules.
\item \textbf{Rearrange Tasks}: ORLA*~\cite{ORLA*} explores object rearrangement tasks in local environments based on \textbf{BestMan}. The diverse and numerous asset supports researchers in testing the generalization of rearrangement algorithms.
\item \textbf{Sim2Real Transfer}: \textbf{BestMan} supports more efficient verification of algorithm deployment based on the unified sim-to-real middleware.
\item \textbf{Multi-agent Collaboration}: MHRC~\cite{MHRC} enables decentralized multi-heterogeneous robot collaboration with LLM based on \textbf{BestMan}. Researchers can efficiently test and optimize the performance of multi-agent systems in collaborative tasks in our \textbf{BestMan} framework.
\end{itemize}

Beyond the aforementioned domains, \textbf{BestMan} demonstrates strong generalizability across mobile manipulation paradigms. Researchers can systematically extend its capabilities through configurable task templates (Table~\ref{tab:platform_comparision}) based on their specific needs.

\section{Conclusions}
In this paper, we introduce \textbf{BestMan}, a scalable real-to-sim-to-real platform designed to enable seamless transfer for household mobile manipulation tasks. 
Our platform offers high-quality articulated scene reconstruction, scalable skill strategy for large-scale simulation testing, and efficient real-world deployments.
We utilize the ASG to generate diverse, fully interactive, and physically plausible scenes.
Furthermore, we construct a modular skill architecture that facilitates flexible algorithmic integration and propose an HUM that supports unified and compatible sim-to-real deployments. Experimental results demonstrate that \textbf{BestMan} outperformed existing methods and provides efficient solutions for complex tasks and environments. Notably, our platform can be applied to a wide range of diverse mobile manipulation tasks in unstructured environments.

\textbf{Funding} This work was supported by the National Natural Science Foundation of China under Grant 62322601.

\section{Declarations}
\textbf{Conflict of interest} The authors have no relevant financial or nonfinancial interests to disclose.

\textbf{Open-source Code} \url{https://github.com/AutonoBot-Lab/BestMan}

\bibliography{references}

\end{document}